\pdfoutput=1

\documentclass[11pt]{article}

\usepackage{acl}
\usepackage[tikz]{bclogo}
\usepackage{lipsum}
\usepackage{times}
\usepackage{latexsym}

\usepackage[T1]{fontenc} 

\usepackage[utf8]{inputenc}

\usepackage{microtype}

\usepackage{booktabs}
\usepackage{graphicx} 
\usepackage{graphics}
\usepackage{amssymb}
\usepackage{amsmath}
\usepackage{pifont}
\usepackage{makecell}
\usepackage{resizegather}
\usepackage{etoolbox}
\usepackage{booktabs,makecell,tabularx} 
\usepackage{enumitem}
\usepackage{mathtools}
\usepackage{cleveref}
\usepackage{multirow}
\usepackage{booktabs,array}
\usepackage{amsmath, bm}
\usepackage{color, colortbl}
\usepackage{xcolor}

\usepackage[ruled,vlined]{algorithm2e}         
\newcolumntype{P}[1]{>{\centering\arraybackslash}p{#1}}
\newcommand{\eqdef}{=\vcentcolon}
\let\oldnl\nl
\definecolor{Gray}{gray}{0.9}
\definecolor{LightCyan}{rgb}{0.88,1,1}
\newcommand{\nonl}{\renewcommand{\nl}{\let\nl\oldnl}}
\newcommand{\comm}[1]{{\nonl{\small{{\color{brown}{/*~#1}~*/}}}}}

\newcolumntype{H}{>{\setbox0=\hbox\bgroup}c<{\egroup}@{}}

\newcommand{\righttriangle}{\mathbin{\rotatebox[origin=c]{30}{$\text{\ding{115}}$}}}

\newcommand{\round}[1]{\ensuremath{\lfloor#1\rceil}}
\newcommand{\T}{$^{\text{(T)}}$}%
\newcommand{\AVG}[1]{$\overline{\text{{#1}}}$}%
\newcommand{\relook}[1]{{#1}}

\title{
\vspace*{-0.5in}
{{\small \hfill ACL 2022}\\
\vspace*{.25in}}
On Continual Model Refinement in Out-of-Distribution Data Streams}

\author{
Bill Yuchen Lin$^\dagger$\thanks{~~The work was done when Bill was an intern at FAIR.} \quad Sida Wang$^\ddagger$\quad Xi Victoria Lin$^\ddagger$ \\  
\textbf{Robin Jia$^\dagger$\quad Lin Xiao$^\ddagger$\quad Xiang Ren$^\dagger$\quad Wen-tau Yih$^\ddagger$}
\\
{
{$^\ddagger$ Facebook AI Research} \quad 
{$^\dagger$}University of Southern California}
\\ 
{\small{\texttt{\{yuchen.lin,robinjia,xiangren\}@usc.edu,\{sida,victorialin,linx,scottyih\}@fb.com }}}
}

\begin{document}
\maketitle

\begin{abstract}

Real-world natural language processing (NLP) models need to be continually updated to fix  
the prediction errors in out-of-distribution (OOD) data streams 
while overcoming catastrophic forgetting. 
However, existing continual learning (CL) problem setups cannot cover such a realistic and complex scenario.
In response to this, we propose a new CL problem formulation dubbed  \textit{continual model refinement} (CMR).
Compared to prior CL settings, CMR is more practical and introduces unique challenges (boundary-agnostic and non-stationary distribution shift, diverse mixtures of multiple OOD data clusters, error-centric streams, etc.).  
We extend several existing CL approaches to the CMR setting and evaluate them extensively. 
For benchmarking and analysis, we propose a general sampling algorithm to obtain dynamic OOD data streams with controllable non-stationarity, as well as a suite of metrics measuring various aspects of online performance.
Our experiments and detailed analysis reveal the promise and challenges of the CMR problem, supporting that studying CMR in dynamic OOD streams can benefit the longevity of deployed NLP models in production. 
\footnote{Our code and data are available at the project website —
\url{https://cmr-nlp.github.io/}.}

\end{abstract}
\section{Introduction}
\label{sec:intro}

{
Fine-tuning large pre-trained language models (LMs) has become the \textit{de facto} standard for training models of a variety of tasks in natural language processing (NLP). 
These success stories are usually in places where the training and testing data are drawn from the same distribution.
However, in real-world scenarios, a \emph{deployed} model (e.g., a question answering service) often encounters examples that are out of the training distribution (i.e., out-of-distribution, \textbf{OOD}).
Such distribution shift often leads to a high error rate. 
In practice, it is highly preferred to continually refine deployed models whenever new errors are reported and annotated, in order to reduce their further negative impacts.
}


{
In spite of its importance, the challenge of continually refining a model over OOD data streams has been underexplored.
Prior work in continual learning (CL) has primarily focused on {task-incremental} settings with {boundary-aware} data streams.
These CL methods are usually evaluated on simple models and data (e.g., image classification with MNIST)~\cite{Aljundi2019OnlineCL}.

It is not clear to what extent they can efficiently refine a model in \textit{boundary-agnostic} streams for a complex language task (e.g., reading comprehension) with modern LMs.
In addition, there is no existing evaluation protocol for comprehensively comparing the collection of applicable methods for such a practical and complex problem. 
Traditional CL paradigms mainly focus on incrementally learning a model from a data stream with a sequence of distinct tasks with explicit delineation, which is rather unrealistic in real-world NLP applications.
}

{
To address these research questions, we propose a novel CL formulation named \textit{continual model refinement} (\textbf{CMR}), which aims to efficiently update a model for error correction in an out-of-distribution data stream without catastrophically forgetting its acquired knowledge over time. 
In contrast to prior CL setups, CMR targets learning a model of a particular task (e.g., question answering) from its prediction errors in dynamic OOD data streams.
Instead of assuming that the streams are drawn from a fixed unseen distribution,
we study CMR under a more general and realistic scenario, 
where the underlying distribution of OOD data streams is \textit{non-stationary} across time steps without clear boundaries while being \textit{diverse} at every time step. 
}

\begin{figure*}[t]
	\centering
	\includegraphics[width=1\linewidth]{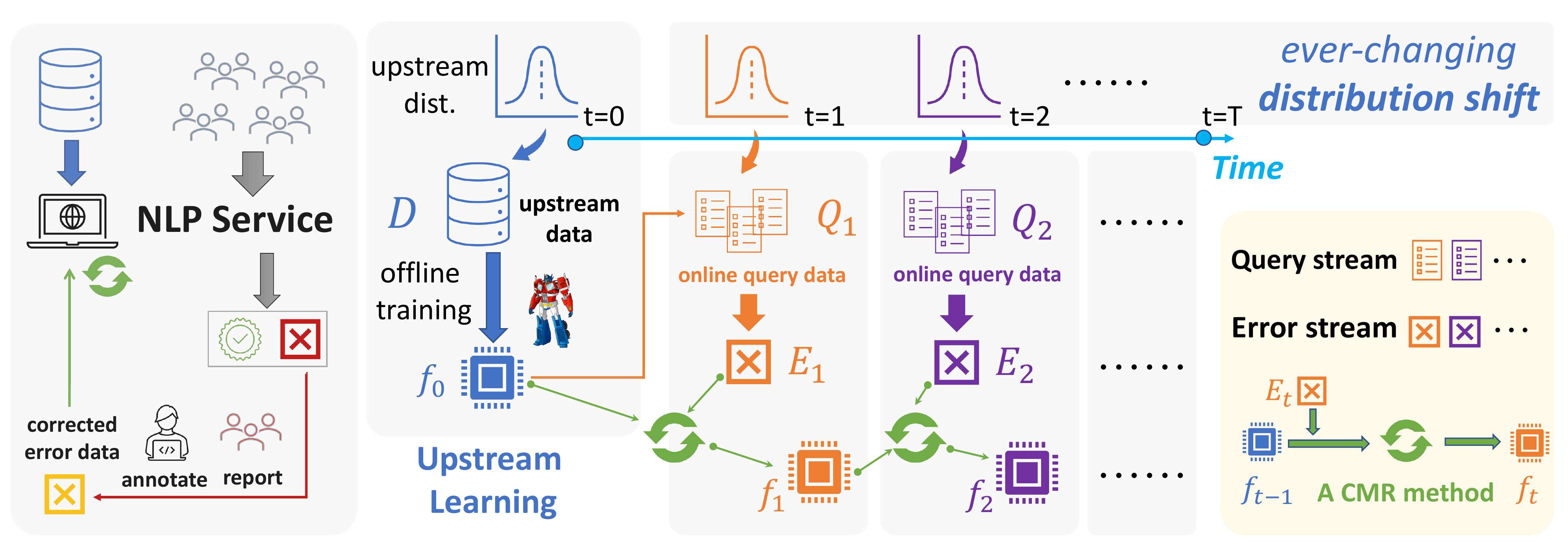}
	\vspace{-1em}
	\caption{\small The continual model refinement (CMR) problem. We offline train a model $f_0$ and it may encounter many error cases $E_t$ when it is tested on a stream of query examples $Q_t$ over time which is drawn from ever-changing unseen distributions.
	A CMR method $g$ aims to fix the error cases over time by refining $f_{t}$ without catastrophic forgetting.   \vspace{-1em}}
	\label{fig:intro}
\end{figure*}


{
In this paper, we focus on studying whether existing methods can address CMR and how we should benchmark and analyze their performance.
We first formulate the CMR problem with several basic metrics covering multiple desiderata for a CMR method: the ability to instantly fix known errors, the retention of previously acquire knowledge from upstream/online data, and the generalization to unseen OOD data (Sec.~\ref{sec:problem}). 
Then, we propose a general method to create the dynamic data streams of the aforementioned characteristics and evaluation metrics to benchmark CMR methods, yielding a comprehensive evaluation protocol for CMR (Sec.~\ref{sec:eval}). 
We employ and extend several suitable methods from the CL literature to study the CMR problem, which is based on \textit{parameter regularization} or \textit{memory replay} (Sec.~\ref{sec:methods}).
}

We have conducted a comprehensive analysis with extensive experimental results, which reveal many interesting, non-trivial findings (Section~\ref{sec:exps}). 
For example, we find that even though replay methods are generally better than regularization-based methods, EWC~\cite{ewc2017}, a typical regularization method, achieves the best score in generalizing to unseen OOD data.
We also find that a simple variant of ranking criteria in conditional replay methods achieves more stable results. Moreover, we find that different CMR methods have orthogonal improvements and our positive initial results suggest that integrating regularization terms for replay methods is a promising future direction to develop advanced CL methods to address CMR.

\begin{figure*}[t]
	\hspace{-1em}
	\includegraphics[width=1.05\linewidth]{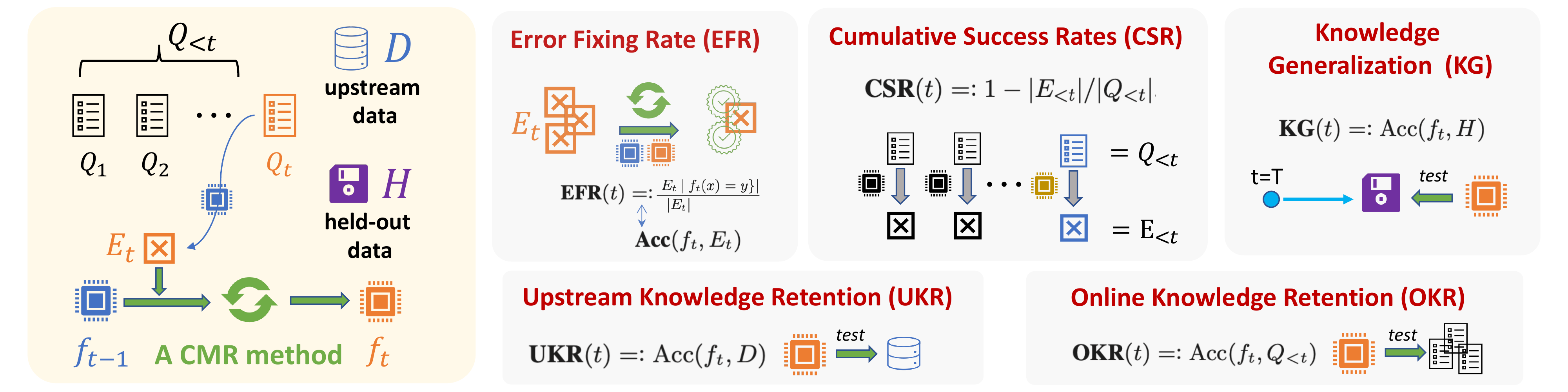}
	\vspace{-1em}
	\caption{ The proposed five basic metrics for evaluating continual model refinement methods. \vspace{-1em}}
	\label{fig:metrics}
\end{figure*}
 
\section{Continual Model Refinement}
\label{sec:problem}
In this section, we formally introduce the proposed continual learning setup, continual model refinement (CMR). 
We first define the notations and describe the learning objectives that are also illustrated in Fig.~\ref{fig:intro}, then we design a few basic evaluation metrics for assessing CMR methods, and finally, we briefly discuss the unique challenges compared to other CL formulations.

\subsection{Problem Formulation}
\label{ssec:problem}

\paragraph{Upstream learning.}
Suppose that we want to build a question answering (QA) model.
To do this, we usually need to offline fine-tune a large pre-trained LM with the existing QA data we have now.
Formally, we denote a dataset with $D=\{(x_i, y_i)\}$, consisting of the examples are drawn from an upstream distribution $\mathcal{U}$, i.e., $D \sim \mathcal{U}$.
The fine-tuned LM is named \textbf{upstream model} $f_0$.

\paragraph{Query streams.}
After the model $f_0$ is deployed in production,  it is common to see ever-changing distribution shifts in real-world data.
We use $\{Q_1, \dots, Q_T\}$ to denote the arriving examples grouped in $T$ episodes and call this sequence of datasets as a \textbf{\textit{query stream}}.
We discuss our method of creating such challenging query streams for evaluating CMR in Sec.~\ref{ssec:nonsta} and Alg.~\ref{alg:cns}.

\paragraph{Error streams.}
In real-world scenarios, the size of $Q_t$ can be very large even in a short period of time, and it is unrealistic to assume that we can annotate all of them to refine the model $f_{t-1}$. 
A common practice is to only annotate the ones that are reported as prediction errors or bugs.
Motivated by this, 
we use $E_{t}$ to denote the examples in $Q_{t}$ that are predicted \textit{incorrectly} by $f_{t-1}$. 
This thus forms an evolving, dynamic stream of prediction errors $\{E_1, \dots, E_T\}$,
where $E_t = \{~(x, y) \in Q_t ~~|~~ f_{t-1}(x) \neq y~\}$.

\paragraph{Learning objectives.}
To improve the user satisfaction over time, we need a \textit{continual model refinement} (CMR) method $g$ that can efficiently take the model $f_{t-1}$ and $E_{t}$ as input and then output a refined model $f_{t}$ for processing future examples.
We expect $f_t$ to output correct answers for the known errors $E_{t}$ immediately while maintaining its correct predictions on previous questions that are answered correctly.
We also want the refined models to keep their generalization ability to unseen future data in the stream.
Sec.~\ref{ssec:metric} shows the metrics to assess a CMR method $g$ toward these goals.


\subsection{Basic Evaluation Metrics}
\label{ssec:metric}
We use five metrics to describe the desiderata for CMR methods and assess them quantitatively, which are illustrated in Figure.~\ref{fig:metrics}. 
We show how to use these metrics for benchmarking in a comprehensive yet concise way in Sec.~\ref{ssec:benchmarking}.




\paragraph{$\bullet$ Error-fixing rates (EFR).}
To assess the responsiveness of the error-fixing methods, 
we look at how many errors can be fixed right away.
We define the instant error-fixing rate at time step $t$ as:
\vspace{-1em}
\par
{{
\small
\begin{align}
\centering
    {
    \operatorname{\textbf{EFR}}(t) \eqdef \operatorname{\textbf{Acc}}(f_t, E_t) \eqdef \frac{|\{(x,y)\in E_t ~|~ f_t(x) = y \}|}{ |E_t|} \nonumber.
    }
\end{align}
}}%




\paragraph{$\bullet$ Knowledge retention (UKR\&OKR).}
We define two metrics below to assess how much knowledge acquired from upstream or online data streams that the model maintains over time:

\vspace{-1em}
\par
{{
\small
\begin{align}
\centering
    {
    \operatorname{\textbf{UKR}}(t)\eqdef\operatorname{Acc}(f_t, D) \text{~~and~~}
 \operatorname{\textbf{OKR}}(t)\eqdef\operatorname{Acc}(f_t, Q_{<t})\nonumber,
    }
\end{align}
}}%
\noindent
where $Q_{<t}$ = $\bigcup^{t-1}_{i=1} Q_i$.
We down-sample $D$ and $Q_{<t}$ and compute periodically for efficiency.


\smallskip

\paragraph{$\bullet$ Cumulative success rates (CSR).}
To monitor the model performance on incoming query examples, 
we compute a running average of success rates at past time steps: $\operatorname{\textbf{CSR}}(t) \eqdef 1- |E_{<t}| / |Q_{<t}|$. This can assess the adaptability of a CMR method.


\smallskip

\paragraph{$\bullet$ Knowledge generalization (KG).}
As we only have a finite number of episodes for experiments, 
to assess the model performance in the future episodes, we  test the models with a \textit{held-out} set of test examples, $H$, that are drawn from the same underlying distributions which are used to create the query stream. 
That is, $\operatorname{\textbf{KG}}(t) \eqdef \operatorname{Acc}(f_t, H)$.


\subsection{Unique Challenges of CMR}
\label{ssec:challenges}

\relook{
Without loss of generality, we suppose that $Q_t \sim \mathcal{O}_t$, where $\{\mathcal{O}_t\}$ denotes an ever-changing series of unseen distributions.
Typical task-incremental CL problem setups such as LAMOL~\cite{Sun2020LAMOLLM} and CLIF~\cite{clif} consider $Q_t$ and $Q_{t+1}$ are sampled from two distinct tasks.
Therefore,
the distribution shifts are sudden (i.e., $\mathcal{O}_t$ and $\mathcal{O}_{t+1}$ does not share any overlapping components).

Also, in conventional CL formulations, the past distribution will never be revisited, which is rather unrealistic in real-world applications.
They do not have the concept of ``error stream'' either.
Instead, the proposed CMR formulation is essentially a boundary-agnostic CL problem in non-stationary data streams, where the distribution shifts are more dynamic, unpredictable, and diverse, yielding a more realistic yet challenging CL setup.
}

\begin{figure*}[th!]
	\centering
	\includegraphics[width=1\linewidth]{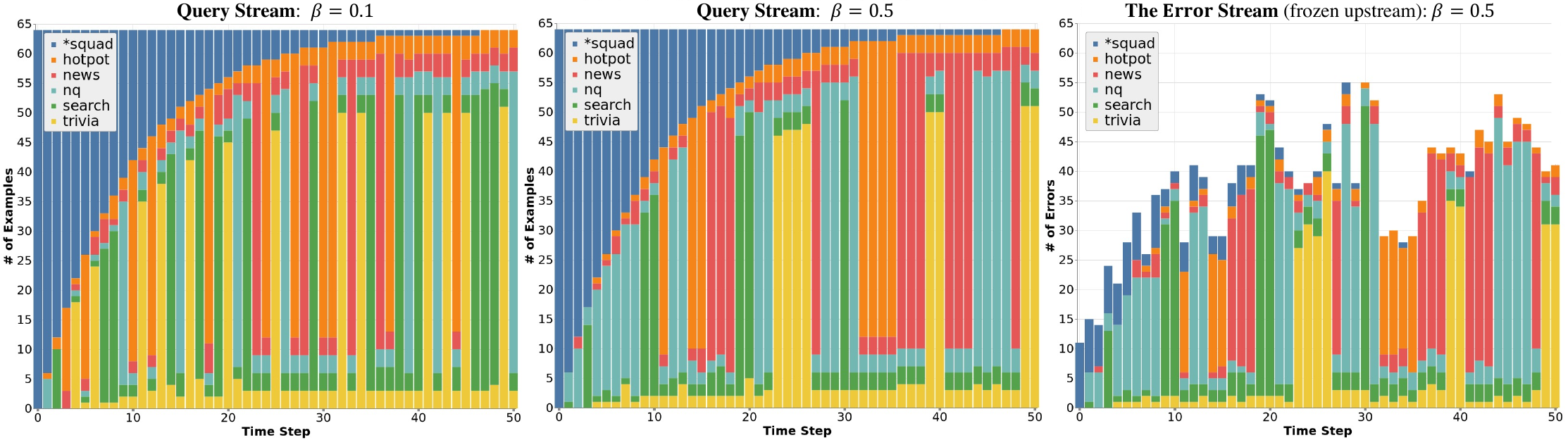}
	\caption{\small The \textit{left} and \textit{middle} figures are two \textbf{query streams} of the QA task with different non-stationarity ($\beta=\{0.1, 0.5\}$) while sharing other arguments ($T$=50, $b$=64, $\alpha$=0.9, $\gamma$=0.8). 
	We use the blue color and `*' to denote the in-distribution data cluster (i.e., $V_0$), the percentage of which decay over time. 
	The distribution of incoming query examples dynamically shifts over time ---
	with larger $\beta$, adjacent episodes are more likely to share the same \textbf{major OOD clusters} which takes the $\gamma$ of the total OOD data.
	To encourage \textit{diversity}, the other $1-\gamma$ OOD examples are sampled from the remaining clusters.
	The \textit{right} figure is the \textbf{error stream} if we do not refine the upstream model (i.e., $f_t\equiv f_0$) and test it on the \textit{middle} query stream. } 
	\label{fig:submission_stream}
\end{figure*}

\section{A Comprehensive Evaluation Protocol}
\label{sec:eval}
We provide a comprehensive evaluation protocol for studying continual model refinement in OOD streams. 
This section first briefly describes our selected task and datasets (Sec.~\ref{ssec:data}), then focuses on our proposed method to sample non-stationary OOD data streams (Sec.~\ref{ssec:nonsta}), and finally, illustrate how we use the basic metrics to benchmark various CMR methods in a comprehensive yet concise way.

\subsection{Datasets}
\label{ssec:data}






In this paper, we mainly use \textit{extractive} question answering (i.e., machine reading comprehension) to evaluate and analyze CMR methods, while one could also study the CMR problem in any NLP tasks with the proposed protocol. 
We use the MRQA-19 benchmark~\cite{fisch-etal-2019-mrqa} which consists of 6 datasets sharing the same formats.

We use the SQuAD~\cite{rajpurkar2016squad10} as the upstream data for offline training the base LM, and use the other five parts as the OOD data for continual learning: NQ~\cite{kwiatkowski2019natural}, HotpotQA~\cite{yang2018hotpotqa}, SearchQA~\cite{Dunn2017SearchQAAN} and TriviaQA~\cite{Trischler2017NewsQAAM}.
This is because SQuAD is more commonly used for deploying models in production and the real-world QA examples from online users can be more similar to the distribution of NQ and SearchQA. 






\subsection{Creating Dynamic OOD Data Streams}
\label{ssec:nonsta}
Here we discuss how to create a realistic ever-changing series of  distributions (i.e., $\{\mathcal{O}_t\}$ in Sec.~\ref{ssec:challenges}) for creating query streams $\{Q_t\}$.

\paragraph{Background.}
A common practice in CL to create a controllable non-stationary data stream is to control the context-switching probability.
For example, OSAKA~\cite{OSAKA}, as a representative method, uses a Markov chain to sample a sequence of tasks with a constant transition probability and then sample the examples from the selected task at each time step.
Despite its simplicity, this method is nevertheless limited to the cases where query stream $Q_t$ can only be drawn from a single distribution, which can be unrealistic. 

Instead, it is common that the online data at a time step are from multiple underlying OOD clusters, each of which has a different distribution, thus yielding a more diverse and challenging environment for continual model refinement. 
Also, it is often that in the early stage of the model deployment, the query streams still contain examples of the upstream distribution~$\mathcal{U}$, and the proportion of such in-distribution examples will decay over time.  


\begin{algorithm}[t]
	\small 
	\nonl \textbf{Input Data Clusters:}  {{$V_0, V_1, \dots, V_N$}} \\
	\nonl \textbf{Configuration Arguments}: $T, b, (\alpha, \beta, \gamma)$.  \\
	\nonl \textbf{Output}: A query stream $\{Q_1, Q_2, \dots, Q_T\}$
	
	 
	\ForEach{$t$ in $\operatorname{range}(1, T)$} { 
	    $b_u = \round{b * \alpha^{t-1}}$; $b_o = b -  b_u$ ; 
	    $b_o' = \round{b_o * \gamma} $ ; \\  
	    $c_t \sim P(c | c_{t-1}; \beta)$ \\
	    \comm{
	    The prob. of switching the major OOD data cluster is $1-\beta$, i.e.,
	    $P(c_t \neq  c_{t-1})=1-\beta$} \\
	    $V_{\neq c_t} = \bigcup_{\{k \in [1,N] | k \neq c_t\}} V_k$ \\ 
	    $Q_t {~\xleftarrow[]{}} \operatorname{sample}(V_0, b_u)$ \\ \comm{$V_0 \sim \mathcal{U}$; from upstream distribution} \\
		$Q_t \text{~+=~} \operatorname{sample}(V_{c_t}, b_o')$ \\ \comm{from the current major OOD data cluster} \\ 
		$Q_t \text{~+=~} \operatorname{sample}(V_{\neq c_t}, b_o-b_o')$ \\ \comm{from non-major data clusters}\\
        {$\operatorname{\textbf{assert}} ~|Q_t| = b$} 
	}
	\caption{\small{Sampling query streams with \textit{controllable non-stationarity} from multiple data clusters.}}\label{alg:cns}
\end{algorithm}

\paragraph{Our proposed method.}
Motivated by these practical considerations, we propose a novel sampling algorithm to control the dynamics of query streams, aiming to encourage diversity and model the decaying upstream distribution.
We consider that there are $N$ underlying data clusters, $\{{V}_1, \dots, {V}_N\}$, each of which corresponds to an unseen distribution, and we have ${V}_0\sim \mathcal{U}$ which is a data set sampled from the upstream distribution.

Our key motivation is to sample the target $Q_t$ from three sources: the in-distribution data cluster $V_0$, the data of a major OOD cluster $V_{c_t}$, and the mix of other remaining OOD data clusters $V_{\neq c_t}$.
As shown in \textbf{Alg.~\ref{alg:cns}},
we have three key configuration arguments $(\alpha, \beta, \gamma)$ for controlling the dynamics of the query stream: 
1) $\alpha$ is the \textbf{decaying factor} for the proportion of in-distribution data, 2) $\beta$ is the \textbf{transition probability} of the Markov chain for deciding the index of the major OOD cluster $c_t$, and 3) $\gamma$ is to control the \textbf{diversity} by adding data from remaining OOD clusters; $T$ is the number of episodes and $b$ is size of $Q_t$.
{Fig.~\ref{fig:submission_stream}} shows examples of query streams and associated error streams.

\begin{figure*}[t]
	\hspace{-0.5em}
	\includegraphics[width=1.03\linewidth]{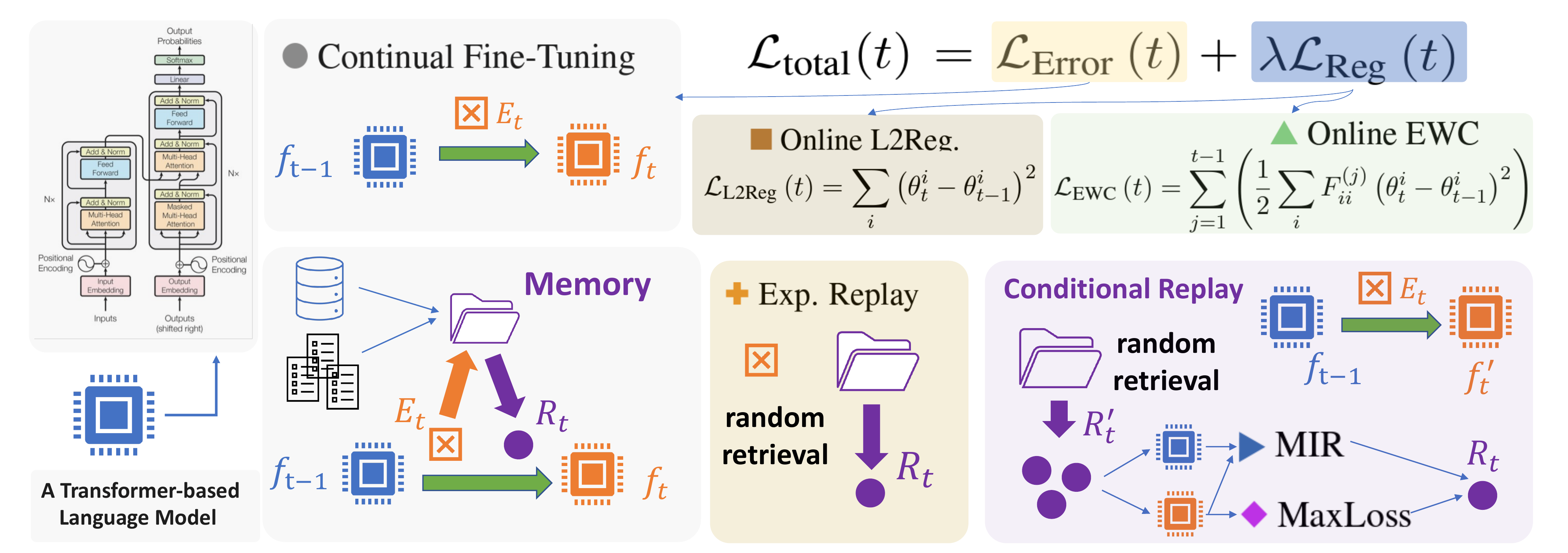}
	\caption{ The studied  methods for continual model refinement (Section~\ref{sec:methods}).  \vspace{-1em}}
	\label{fig:methods}
\end{figure*}

\subsection{Benchmarking CMR Methods}
\label{ssec:benchmarking}
\paragraph{Overall measurement.}
Recall that there are five basic metrics in Section~\ref{ssec:metric}, namely \textbf{EFR} (instnat error-fixing rate), \textbf{UKR} (upstream knowledge retention), \textbf{OKR} (online knowledge retention), \textbf{CSR} (cumulative success rate) and \textbf{KG} (knowledge generalization). 
To have a comprehensive yet concise analysis of CMR methods, 
we report the average and final values of these metrics. 
Specifically, we use \AVG{X} to denote the average scores in the metric X (e.g., X=UKR) over all time steps, and X\T~to denote the score at the final time step.
Reporting both can help us quickly assess the trend of performance of $f_t$ in addition to its final performance.
Besides these fine-grained scores, we also provide an \textit{overall evaluation criterion} (OEC) by taking the average of the four scores except for the EFRs\footnote{{Note that we report \AVG{EFR} scores separately because it computes on the method-specific errors unlike other metrics that test on same examples for all CMR methods.} 
}, i.e., $\text{OEC} = \operatorname{average}(\text{UKR, OKR, CSR, KG})$.

{
\paragraph{Validation/testing streams.}
To evaluate CMR methods (introduced later in Sec.~\ref{sec:methods}), 
we use the method in Alg.~\ref{alg:cns} to sample multiple streams under the same configurations (i.e., $T,b,\alpha, \beta, \gamma$ and $\{V_i\}$) and then split them as validation streams and testing streams.
The validation streams are used to pick the best hyper-parameters of each CMR method (e.g., the $\lambda$ of regularization-based methods and the size of $R_t$ in replay methods) and then they are evaluated on the same set of testing streams.
}

\section{Methods}
\label{sec:methods}

We first introduce our base LM and then illustrate several typical continual learning methods with our extensions to make them applicable to the CMR problem. We discuss other relevant yet not suitable methods in \textit{Related Work} (Sec.~\ref{sec:related}). 
We use Figure~\ref{fig:methods} to illustrate these CMR methods.

\subsection{Base Model \& Continual Fine-Tuning} 
\paragraph{Base model.}

Pretrained text-to-text language models, such as BART~\cite{lewis2019bart} and T5~\cite{t5}, are commonly used for studying a wide range of NLP tasks. 
This is because they are generally applicable to tasks that can be formulated as a text-to-text problem, and that they show better generalization potential~\cite{ye-etal-2021-crossfit, Wei2021FinetunedLM, sanh2021t0}.
We thus employ the text-to-text formats to pre-process all data in our experiments and use \textit{BART-base} as the base model.
We find the BART-base model is a great fit to support our extensive experiments for its relatively smaller size and comparable upstream performance versus its alternatives. 
Thus, we use it for our experiments to ensure the scalability of our analysis and the generality of our findings. 
Note that we do not aim to offline train a perfect upstream model $f_0$ with the upstream dataset $D$. 
Instead, we focus on the CMR methods that can continually refine a given upstream model.

\paragraph{Continual fine-tuning.}
The most straightforward method is to always use a vanilla optimizer (e.g., Adam~\cite{kingma2015adamam}) to fine-tune $f_{t-1}$ with a small learning rate on $E_{t}$ for a few epochs, aiming to minimize the loss $L_{\text{Error}}(t)$ of fine-tuned model $f_t$ on $E_t$.
Such refined models $f_{t}$ should be able to output correct outputs for these known errors. 
This method may overfit these errors and thus forget previously acquired knowledge.
We introduce a few regularization methods next.

\subsection{Regularization-based methods}
A common solution to preventing forgetting is to add a temporal regularization term to the loss for continual fine-tuning:
$
    \mathcal{L}_{\text{total}}(t)=\mathcal{L}_{\text {Error }}(t)+\lambda \mathcal{L}_{\text {Reg }}(t)
$, so that the parameter changes from $f_{t-1}$ to $f_{t}$  are restricted to avoid over-fitting.

\paragraph{Online L2Reg.}
We use an intuitive regularization term by computing the L2 distance between the parameters. That is, $$\mathcal{L}_{\text {L2Reg }}(t) = \sum_{i} \left(\theta_{t}^{i}-{\theta}_{t-1}^{i}\right)^{2},$$ where $\theta_{t}$ is the parameters of $f_t$. This regularization term mitigates the forgetting issue by applying a penalty for every parameter change.

\paragraph{Online EWC.} Elastic weight consolidation~\cite{ewc2017} is a typical regularization method for CL. 
Unlike L2Reg which gives an equal penalty to every parameter change, 
EWC produces a weighted penalty such that the parameters that are more important to the previous tasks will have larger penalty weights, leading the parameter changes to find an overlapping space where both previous knowledge and new knowledge can be stored in the parameters.
In particular, it efficiently estimates the Fisher Information Matrices ${F}_{i i}^{(t)}$  and use them for consolidating the weighted penalty:
$$\mathcal{L}_{\text {EWC }}(t) = \sum_{j=1}^{t-1} \left( \frac{1}{2} \sum_{i} {F}_{i i}^{(j)} \left(\theta_{t}^{i}-{\theta}_{t-1}^{i}\right)^{2} \right).$$ 
We here employ an extension of EWC by keeping a running sum of $F_{ii}$ to avoid the growth of computation cost in the online setting. 


\subsection{Replay Methods}
The other significant group of CL methods is based on replaying past examples, as follows:

\paragraph{Experience replay.} ER~\cite{Rolnick2019ExperienceRF} is a simple yet effective replay method that stores the previous examples into a growing memory module $M$.
Then, we periodically (every $k$ time steps) sample a small subset of the memory $R_t$ as additional training examples for model refinement.
It uses a two-stage process: fine-tune $f_{t-1}$ on $R_t$ to get $f_{t-1}'$ and then fine-tune $f_{t-1}'$ on $E_t$ to get $f_t$.

\paragraph{Maximally interfered replay (MIR).}
Instead of randomly selecting $R_t$ from $M$, {MIR}~\cite{Aljundi2019OnlineCL} aims to replay the \textit{most forgettable} examples, conditioning on the current information: $f_{t-1}$ and $E_t$.
It samples a small candidate pool $C\subset M$ and then ranks the examples in $C$ by their ``\textit{interference scores}.'' Finally, the $R_t$ of MIR is the subset of $C$ with the largest scores.
To compute interference scores, we first fine-tune $f_{t-1}$ on $E_t$ to get a \textbf{virtual model} $\hat{f_t}$. 
Then, we compute the loss of $f_{t-1}$ and $\hat{f_t}$ on each example in $C$ to get the interference scores (i.e., the loss delta): $$\operatorname{score}(x_i,y_i) \eqdef \operatorname{loss}(\hat{f_t}(x_i), y_i) -  \operatorname{loss}(f_{t-1}(x_i), y_i).$$


\paragraph{MaxLoss replay.} 
Inspired by~\citet{Jiang2019AcceleratingDL} and~\citet{Kawaguchi2020OrderedSA} that show learning with the examples with largest losses can enhance the learning efficiency, we propose a variant of the MIR by redefining the scoring function to $\operatorname{score}'(x_i,y_i) \eqdef \operatorname{loss}(\hat{f_t}(x_i), y_i)$ and call it MaxLoss, which takes the examples that have largest losses on the virtual model $\hat{f_t}$ (instead of the largest delta in MIR).
 
\paragraph{Extension for CMR.} 
(1) \textbf{Bi-Memory:} There are two types of  knowledge that we want to maintain in CMR: the knowledge acquired in upstream and online learning respectively. 
Considering that the upstream data is much larger than the incoming errors, it is thus not reasonable to use a single memory module as in other CL problems.
We thus use two separate memory modules $M_u$ and $M_o$ where the upstream memory is $M_u=D$ and the online memory $M_o$ grows by adding $E_t$.
(2) \textbf{Mixed-Tuning}: Instead of following the two-stage method of using $R_t$,
we choose to mix $R_t$ and $E_t$ for fine-tuning $f_{t-1}$.
Both modifications are supported by their better empirical results.






\begin{table*}[th!]
\centering
\scalebox{0.75}{
\begin{tabular}{r||c||cccc|c||cccc|c}
\toprule
  \textbf{Methods} $\downarrow$ \textbf{Metrics} $\rightarrow$ & \cellcolor{red!10} \AVG{EFR} & \cellcolor{blue!10} \AVG{UKR} & \cellcolor{blue!10} \AVG{OKR} & \cellcolor{blue!10} \AVG{CSR} & \cellcolor{blue!10} \AVG{KG} & \cellcolor{blue!10} \textbf{\AVG{OEC}} & \cellcolor{cyan!10} UKR\T & \cellcolor{cyan!10} OKR\T & \cellcolor{cyan!10} CSR\T & \cellcolor{cyan!10}KG\T & \cellcolor{cyan!10} \textbf{OEC\T} \\ \midrule 

 \rowcolor{gray!10}  Frozen Upstream ($f_t \equiv f_0$)  & 0.00  & 80.27 & 43.69 & 44.95 & 31.25 & 50.04 & 80.27 & 36.13 & 35.44 & 31.25 & 45.77 \\ \midrule
 {\color{gray}\ding{108}} \underline{Continual Fine-Tuning}   & 97.36 & 72.05 & 83.87 & 55.93 & 45.68 & 64.38 & 66.21 & 77.73 & 53.48 & 48.91 & 61.58 \\  \midrule 
{\color{brown}\ding{110}} \underline{Online L2Reg. }         & 97.18 & 73.47 & 85.37 & 57.27 & 47.12 & 65.81 & 71.09 & 83.59 & 54.50 & 51.17 & 65.09 \\
{\color[HTML]{7fc97f} \ding{115}} \underline{Online EWC}         & 97.49 & 73.38 & 86.09 & 56.17 & 47.34 & 65.75 & 68.55 & 85.74 & 53.67 & 53.28 & 65.31 \\ \midrule
{\color[HTML]{eb9634}\ding{58}} \underline{Exp. Replay (k=3)}    & 97.07 & 75.30 & 87.29 & 56.02 & 47.61 & 66.55 & 72.46 & 87.30 & 54.08 & 52.66 & 66.63 \\
Experience Replay (k=1)    & 96.72 & 78.91 & 89.38 & 57.80 & 47.17 & 68.31 & 78.13 & 86.52 & 55.33 & 52.73 & 68.18 \\ \midrule
{\color[HTML]{c334eb} \ding{117}} \underline{MaxLoss (k=3,c=256)}      & 97.43 & 75.43 & 86.89 & 57.14 & 46.70 & 66.54 & 75.00 & 84.77 & 55.11 & 51.33 & 66.55 \\
MaxLoss (k=1,c=256)      & 96.54 & 78.16 & 89.86 & 57.78 & 46.63 & 68.11 & 77.54 & 89.26 & 55.47 & 50.94 & 68.30 \\
MaxLoss (k=1,c=512)       & 97.41 & 75.57 & 87.09 & 56.80 & 46.45 & 66.48 & 77.54 & 89.65 & 55.88 & 52.81 & 68.97 \\
MaxLoss (k=1,c=1024)   & 96.63 & 77.61 & 89.82 & 58.13 & 47.10 & 68.17 & 80.66 & 91.02 & 55.88 & 50.78 & 69.59 \\
 \midrule
{\color[HTML]{386cb0}$\righttriangle$} \underline{MIR (k=3,c=256)} & 97.08 & 75.92 & 87.13 & 56.91 & 47.22 & 66.79 & 75.78 & 87.50 & 54.53 & 51.80 & 67.40 \\
MIR (k=1,c=256)     & 96.59 & 77.84 & 89.77 & 58.35 & 47.28 & 68.31 & 79.49 & 90.43 & 55.91 & 51.25 & 69.27 \\
MIR (k=1,c=512)     & 96.96 & 77.86 & 89.41 & 58.13 & 46.40 & 67.95 & 79.69 & 89.45 & 55.50 & 50.08 & 68.68 \\
MIR (k=1,c=1024) & 96.71&	77.47&	87.83&	57.98&	46.87&	67.54&	78.13&	87.89&	55.73&	50.70&	68.11 \\ \midrule
{MIR(1,256)}+OnlineL2Reg          & 96.15&	79.10&	90.41&	59.80&	47.90&	69.30 & 79.49 & 90.04 & 57.45 & 52.66 & 69.91 \\
\midrule
 \rowcolor{gray!20}   Offline Refining ($f_0 \rightarrow f_T$)    & \textit{95.62} &	- & - & - &	- & - & \textit{83.78} & 	\textit{93.75} & \textit{93.81} &	\textit{56.17} &	\textit{81.88 }
 \\
\bottomrule

\end{tabular}
}
\caption{\small Results (\%) in multiple metrics: \textbf{EFR}=Error-Fixing Rate; \textbf{UKR/OKR}=Upstream/Online Knowledge Retention; \textbf{CSR}=Cumulative Success Rate; \textbf{KG}=Knowledge Generalization. \textbf{OEC} is the average of the last four. Column names with bars are the average of all periods. The ones with `\T' are the scores at the final step.
The underlined methods are matched with the legends in Figure~\ref{fig:curves}. $k$ is the replay interval (the smaller the more frequent), and $c$ is the size of the candidate pool.  }
\label{tab:main}
\end{table*}

\begin{figure*}[th!]
	\centering
	\includegraphics[width=1\linewidth]{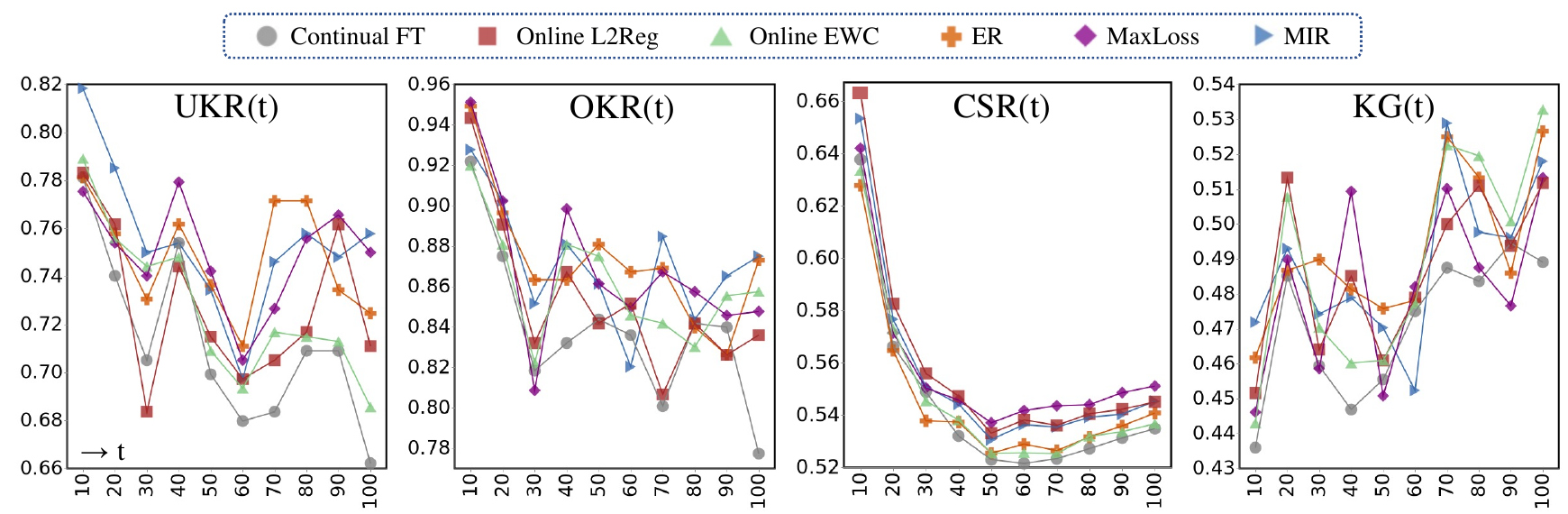}
	\caption{The curves of four key metrics over time of selected CMR methods in Table~\ref{tab:main}.  The $x$-axis is the time step.}
	\label{fig:curves}
\end{figure*}

\section{Evaluation \& Analysis}
\label{sec:exps}

We first present the setup in Sec.~\ref{ssec:evalsetup}, and report our main results in Table~\ref{tab:main} and Figure~\ref{fig:curves}, which we use to discuss our key findings in Sec.~\ref{ssec:mainfindings}~to~\ref{ssec:additionalanalysis}.

\subsection{Setup}
\label{ssec:evalsetup}

\paragraph{Reference range.}

To get a reference range of the performance, 
we set up two reference methods.
1) \textit{\textbf{FrozenUpstream}}: We always use the upstream model (i.e., $f_t \equiv f_0$) for inference at every time step. 
2) \textit{\textbf{OfflineRefining}}: We combine all the errors of $f_0$ as $E_{\leq T}$ and then offline fine-tune the model $f_0$ with $D'+E_{\leq T}$, where $D'$ is a subset of $D$, to directly get the final refined model $f_T$ . 

\paragraph{Hyper-parameters.}  We here use a normal configuration of the streams (i.e., $T$=100, $b$=64, $\alpha$=0.9, $\beta$=0.5, $\gamma$=0.8) for studying the CMR methods and discuss other extreme configurations briefly in Sec.~\ref{ssec:additionalanalysis} and more in Appendix. To select the optimal hyper-parameters of each method (e.g., the learning rate, training epochs, method-specific arguments, etc.), we use grid search and pick the ones with the best overall score on validation streams.


\subsection{Main Results and Findings}
\label{ssec:mainfindings}
We report the results in Table~\ref{tab:main} \& Figure~\ref{fig:curves},
and organize our findings by answering a coherent list of analysis questions: \textbf{\texttt{(Q1-Q7)}}.

\paragraph{\texttt{(Q1)} Can we fix errors without forgetting?} 
From the \AVG{EFR} column, we can see that all methods can achieve a 95+\% instant error-fixing rate, meaning that they can indeed quickly fix most of the known errors. 
However, they tend to forget the previously fixed errors and even examples that are correctly predicted before in the query stream. 
An oracle method that does not forget the previously acquired knowledge would have an OKR\T~of nearly 100\%, while the OKR\T~of the continual fine-tuning method is only $77.7\%$.
The issue of forgetting both online and upstream knowledge in the \textit{continual fine-tuning} baseline is quite serious.
Notably, its OKR\T{} is much lower than its \AVG{OKR} (83.87$\rightarrow$77.73), and similarly for UKR\T{} and \AVG{UKR} (72.05$\rightarrow$66.21).
The curves in Figure~\ref{fig:curves} also suggest that the forgetting issue can be increasingly more serious over time, and it does not show any trend to diminish after $T$. 
\textit{This confirms that studying the CMR problem is of great importance for enhancing deployed NLP models.}


\paragraph{\texttt{(Q2)} How well do CMR methods mitigate the forgetting issue?} 
All tested CMR methods can indeed mitigate forgetting without lowering down the EFRs, but they behave quite differently.
The regularization methods (i.e., Online L2Reg and Online EWC) are better at improving OKRs rather than UKRs, while replay methods enhance both OKRs and UKRs quite well. For example, MaxLoss can achieve the best OKR\T ($91.0\%$) while having a UKR\T that is even slightly better than the FrozenUpstream model (80.6 vs 80.3).

Moreover, we find that MaxLoss and MIR have great potential to continually improve knowledge retention in the future.
From both curves in Fig.~\ref{fig:curves} and Table~\ref{tab:main} (i.e., the comparisons between \AVG{UKR}/\AVG{OKR} and UKR\T/OKR\T), we can see they tend to have better scores in the later stages, but the retention scores of regularization-based methods are decreasing over time. 
We have a detailed discussion on replay-based methods in \texttt{Q4}.

\paragraph{\texttt{(Q3)} Can refined models generalize to unseen OOD data?}
Recall that CSRs evaluate the incoming yet \textit{not} touched examples over time in the stream and the KGs evaluate the held-out examples that are \textit{not} in the stream. 
Both metrics thus test on OOD examples that are unseen to the refined model at that time.
Compared to the \textit{FrozenUpstream} baseline, we see all methods have large performance gains (from 30\% to 50+\% in CSR\T and KG\T).
The ``MIR w/ Online L2Reg'' even achieves the best CSR\T~and it is significantly better than others, showing that learning with replay effectively improves the generalization ability.

From the \AVG{KG} and KG\T~columns of these CMR methods (and Fig.~\ref{fig:curves}), 
we can see that refined models are increasingly more generalizable to held-out unseen data over time as well.
However, the differences among these methods in these two metrics are not obvious, although they are all better than the continual fine-tuning baseline.
Interestingly, the regularization method OnlineEWC gets the best score of KG\T, even though its CSR\T~is worse than others.
\textit{This suggests that learning with replay might hurt the held-out knowledge generalization, but regularization could maintain a better generalization ability in the long run.}


\subsection{Analysis on Memory Replaying}
\label{ssec:memoryanalysis}

\paragraph{\texttt{(Q4)} How should we replay the memory?} 
We find that increasing the replay frequency (i.e., setting a smaller replay interval $k$) can largely improve the overall performance for ER, MaxLoss, and MIR. 
This is expected as there are more fine-tuning steps over the retrieved data.

However, the reason for such improvement varies among them.
Increasing the replay frequency primarily benefits ER's UKR\T, but not for other metrics, and it even causes a lower OKR\T.
Instead, MaxLoss and MIR also benefit from larger OKR\T~(MaxLoss: 84.77 $\rightarrow$ 89.26; MIR: 87.50 $\rightarrow$ 90.43).
\textit{This suggests that conditional replay methods can get more important stored memory to replay than ER's \textit{random} selections. }
Thus, it is promising to develop more advanced \textit{conditional} replay methods for CMR.

\paragraph{\texttt{(Q5)} Are larger buffer sizes always better for conditional replay methods?}
Larger buffer sizes (i.e., c=256 $\rightarrow$ 512 $\rightarrow$ 1024) can increase MaxLoss's UKR\T~and OKR\T with a large margin and thus produce better overall scores. 
However, MIR with larger buffer sizes suffers from decreasing UKR\T~and OKR\T.
This indicates that that \textit{delta} of loss as the ranking criteria is less stable than using the virtual loss itself (i.e., MaxLoss).

{This finding conflicts with the MIR experiments on MNIST-based task-aware streams~\cite{Aljundi2019OnlineCL}.}
We thus conjecture it is because \textit{our streams are more complex and the loss landscapes of the task are significantly different from the toy datasets used for evaluation in many prior CL works} (e.g., image classification over shuffled MNIST).



\begin{figure}[t]
	\centering
	\includegraphics[width=1\linewidth]{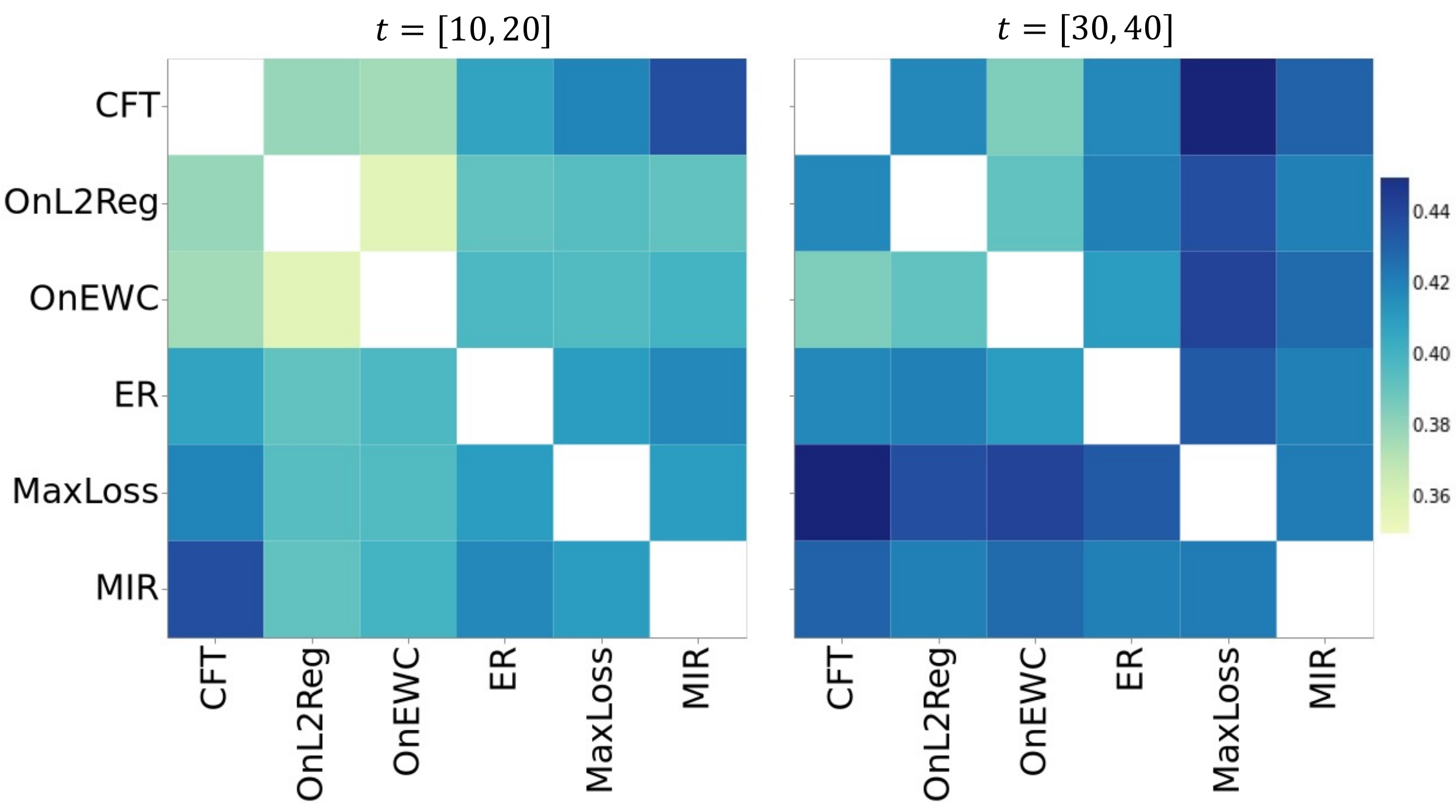}
	\caption{  The differences between refined models produced by different CMR methods in terms of their predictions for the same inputs at two time spans (10-20 and 30-40). The darker cells have large discrepancy.}
	\label{fig:heatmap}
\end{figure}

\subsection{\relook{Orthogonal Improvement for CMR}}
\label{ssec:orthogonal}

\paragraph{\texttt{(Q6)} Do different CMR methods produce similar refined models?}
We use Figure~\ref{fig:heatmap} to visualize the differences among the refined models produced by selected CMR methods in two different periods.
We can see the refined models by continual fine-tuning (CFT) and regularization methods are more similar to each other, and all replay methods are quite distinct from other methods. 
Also, the divergence among different methods rapidly increases from $t=[10, 20]$ to $t=[30, 40]$. 
\textit{Therefore, we believe that the improvement of these CMR methods is orthogonal to each other, especially between regularization and replay methods.}

\paragraph{\texttt{(Q7)} Can we integrate regularization and replay methods?}
Inspired by Fig.~\ref{fig:heatmap} and findings in \texttt{(Q3)}, we add an initial experiment by combining the \textit{MIR} and \textit{OnlineL2Reg} and show its performance in Table~\ref{tab:main}.
Interestingly, we indeed observe this combination produces a noticeable improvement over both \textit{MIR} and \textit{OnlineL2Reg}, yielding the state-of-the-art performance in OEC\T~scores. 
To the best of our knowledge, there is little prior work that has studied the effect of integrating regularization in (conditional) replay methods, and \textit{our initial results suggest that this is a very promising direction for future research}.





\subsection{Additional Analysis}
\label{ssec:additionalanalysis}


Our above analysis is based on the results of a normal stream configuration (i.e., $\alpha$=0.9, $\beta$=0.5, $\gamma$=0.8), but can such tuned hyper-parameters of CMR methods directly apply to streams of extreme configurations? 
In Table~\ref{tab:settings}, we briefly compare the gain of the previous CMR methods in terms of their OCE\T~ improvement over the vanilla FrozenUpstream baseline under a few extreme settings of 
We find that, in general, all replay methods are still better than continual fine-tuning and Online EWC.
ER shows more stable results in extreme settings (e.g., $\beta$= 0.1 or  0.9) but MIR and MaxLoss (MxLs) are more sensitive to the non-stationarity yet less sensitive to the diversity.








\begin{table}[t]
\centering
\scalebox{0.74}{
\begin{tabular}{r||ccccc}
\toprule 
 \textbf{Stream Dynamics}                              & \textbf{CFT}     & \textbf{EWC}   & \textbf{ER}      & \textbf{MxLs}   & \textbf{MIR}   \\ \toprule
 
 
\rowcolor{gray!20} 
 $\alpha$=0.9, $\beta$=0.5, $\gamma$=0.8 & 15.81&	19.54&	20.86&	20.78&	21.63   \\ \midrule
 $\alpha$=0.9, $\beta$=\textbf{0.1}, $\gamma$=0.8 & 23.40&	24.14&	26.32&	26.05&	26.04 \\
 $\alpha$=0.9, $\beta$=\textbf{0.9}, $\gamma$=0.8 & 18.61&	19.38&	20.78&	19.51&	20.60 \\
 \midrule
 $\alpha$=0.9, $\beta$=0.5, $\gamma$=\textbf{0.5} & 19.97&	20.10&	21.97&	23.01&	22.04 \\
 $\alpha$=0.9, $\beta$=0.5, $\gamma$=\textbf{0.2} & 17.37&	16.22&	19.15&	20.60&	19.45 \\
 \bottomrule

\end{tabular} 
}
\caption{\small The \textbf{\textit{gain}} of OEC\T over the \textit{Frozen Upstream} baseline for each method under different stream dynamics. \vspace{-0.5em}}
\label{tab:settings}
\end{table}

\section{Related Work}
\label{sec:related}

\paragraph{Continual Learning for NLP.}
Recently, continual learning (or lifelong learning) has drawn attention in the NLP field~\cite{biesialska2020continual, Sun2020LAMOLLM, wang2019sentence, huang2021continual, clif}.
However, most of these works follow the traditional \textit{task-incremental}, \textit{boundary-aware}, \textit{never-revisiting} CL setup, which is not directly beneficial to most of the real-world scenarios of deployed NLP models.
For example, the CLIF formulation~\cite{clif} focuses on learning over a sequence of different NLP tasks with few-shot data so that the trained model can generalize better to unseen tasks.
In contrast, the proposed CMR in this work is a particularly novel CL setup where we focus on continually refining a model with its prediction errors in OOD data streams, thus yielding a boundary-agnostic, dynamically non-stationary environment for CL methods to work. 
Such fundamental differences between CMR and traditional CL setups make it difficult to directly apply many CL methods that are based on boundary-aware streams, especially for those who require learning task representations.

\paragraph{CMR vs. OSAKA}
The OSAKA~\cite{OSAKA} problem is similar to the CMR in that we both focus on CL in non-stationary boundary-agnostic data streams. However, it does not consider the distribution diversity inside each time step or the decay of upstream distribution in the online setting.
Our sampling method (Alg.~\ref{alg:cns}) fills the gap and yields a more realistic CL setup.
In addition, the data streams of CMR are always the prediction errors of the latest model, thus producing a naturally evolving and adversarial environment for CL methods to explore. 
Moreover, the experiments of OSAKA are limited to simple networks and tasks such as MNIST, but our work uses pre-trained Transformer LMs and the QA task, and thus we believe our analysis and findings are more useful for the NLP community and beyond.

\paragraph{Model Refinement.}
Model refinement has recently become an emerging topic in NLP, but existing works have mainly been limited to  \textit{offline} editing time-sensitive factual knowledge in pre-trained LMs~\cite{zhu2020modifying, de2021editing, Mitchell2021FastME}.
In contrast, our work studies the model refinement in an online continual learning setting and for downstream NLP tasks such as reading comprehension and natural language inference.
\citet{Jang2021TowardsCK} attempt to study the knowledge editing problem at a larger scale, but its problem formulation only contains two time-steps, thus being significantly different from CMR.
\citet{dhingra2021time} propose a simple method to jointly model text with its timestamp so that the trained language models can be calibrated when new knowledge arrives, while CMR focuses on the error cases from OOD data streams where the timestamps have little correlation with the skills we want the deployed model to learn.
Besides, \citet{Yao2021RefiningNN} propose a method of learning from explanations to fix prediction errors, which shares similar high-level motivation but has few direct connections to our focus in this work.

\section{Conclusion \& Future Directions}
\label{sec:conclusion} 

{
In this paper, we propose a novel continual learning formulation named continual model refinement (CMR). The CMR problem aims to efficiently fix prediction errors when learning in out-of-distribution data streams without catastrophically forgetting the acquired knowledge. 
For studying such a realistic and complex problem, 
we presented a dedicated evaluation protocol with a general method to create non-stationary, diverse OOD data streams for analysis.
Also, we design multiple evaluation metrics to deliver a comprehensive yet concise measurement of CMR methods.

The proposed CMR problem with our comprehensive analysis opens up a range of new opportunities for studying continual learning problems that are closer to real-world applications for the NLP community and beyond. 
For example, based on our results and analysis about \texttt{(Q3)} and \texttt{(Q6)}, we find that it is promising to study how we can integrate both regularization methods and replay methods for mitigating the forgetting issue while improving the generalization ability.
The analysis about \texttt{(Q5)} suggests that developing more stable ranking criteria is also important to conditional replay methods (e.g., our simple extension MaxLoss can outperform MIR under specific settings).
Developing CMR methods of which the configurations can generalize to diverse types of streams is also an important challenge. 
We release our codebase and processed datasets for supporting the reproducibility of our experiments and future research. 
}


\bibliography{custom}
\bibliographystyle{acl_natbib}

\clearpage
\appendix

\section{Implementation Details}

\subsection{Upstream Learning}
We use the huggingface's implementation of Transformer architectures for running BartForCondtionalGeneration.
Note that we choose to use this seq2seq head instead of the BartForQuestionAnswering for the seq2seq version can support a much wider range of NLP tasks as long as they can be converted in to text-to-text formats (e.g,. CrossFit~\cite{ye-etal-2021-crossfit} and FLAN~\cite{Wei2021FinetunedLM}).
Also, we find that the results using  seq2seq formats is comparable to using the span extraction for reading comprehension (at least for SQuAD). 
Therefore, we choose to use seq2seq format to encourage the generality of our released codebase.
We use BART-base for all our experiments and here we present the final hyper-parameters we used for upstream learning:  lr=5e-5, train\_bsz=64, pred\_bsz=64, num\_epochs=30. 

We have also tried to use \textbf{BART-Large} for running our experiments and analysis. Our preliminary results show that our general findings still hold such as Q1 to Q3. 
But running BART-Large causes around 5 times slower speed for our experiments.
Considering the scale of our grid search and our analysis as well as the negative impact to the environment, we choose to focus on using BART-base for all our experiments and analysis. 
We believe future works for CMR can also benefit from this due to the fact that using BART-base can help them quickly analyze the performance.
Also, as we seek to test different CMR methods instead of different base LMs, we think using BART-base can represent a reasonable scope of similar LMs that are widely used in the community such as T5-base, etc.

\subsection{Details for CMR Methods}

\paragraph{Datasets.}
We refer to the MRQA 2019 homepage for more detailed statistics of each dataset:  \url{https://github.com/mrqa/MRQA-Shared-Task-2019}.
Particularly, we use the SQuAD-train as $D$, the upstream data, and SQuAD-dev as $V_0$ (the upstream data cluster); Also, other devs as $\{V_1, \dots, V_N\}$ as the OOD data clusters.
We also tried to use NQ as the upstream data and it shows a similar performance trend as we discussed in the main table.

\paragraph{Validation/Test Streams}
We sampled {32 validation streams and 8 test streams} for all our experiments, shown in Table~\ref{tab:main}.
We searched the hyperparameters (hps) of all CMR methods on the set of validation streams and then pick the best one for each method by measuring their average of OEC\T and EFR\T on the sum of all validation streams. 
The results are based on the average of all test streams, where for each stream we run each method with \textbf{5 different random seeds}, yielding 40 rounds of experiments for each CMR method (a reason why we choose to use BART-base).

\paragraph{Continual Fine-Tuning}
There are two major hps: the learning rate and the num\_epochs, we searched over $\{1e-5, 2e-5, \dots, 5e-5\}$ and $\{5, 10, 15, 20, 30\}$ for the num\_epochs at each episode. 
We use the mini-batch size of $8$ for fine-tuning the $f_{t-1}$ on $E_t$ at each time step.
Our final choices are lr=3e-5 and num\_epochs=20. 

\paragraph{OnlineL2Reg}
There is one additional hp: the $\lambda$, the weight of the L2 penalty. 
We searched it from $\{1, 5, 10, 20, 50, 100\}$ on top of the hps of the CFT and finally decide to use $\lambda=10$.

\paragraph{OnlineEWC}
Please refer to the original paper~\cite{ewc2017} for the details of the online version.
Therefore, we also have two hps $\lambda_{\text{EWC}}$ and $\gamma_{\text{EWC}}$, which we searched over $\{1, 5, 10\}$ and $\{1.0, 0.95, 0.9, 0.8\}$.
We finally use $5$ and $0.9$ for their best performance.

\paragraph{Replay Methods}
For all replay methods, we first search them with the best hps using CFT and then run them together with the same size of replay examples $|R_t|=32$ which we found perform the best. 
Their $k$ and $c$ are compared in Table~\ref{tab:main}.

We leave more details of the MIR and MaxLoss implementation in our codebase.

\subsection{Computational Cost}
Replay methods (with best searched hps) are slightly more expensive than continual learning methods. 
Online L2Reg needs to store the weight of the previous model checkpoint and compute the L2 distance, and OnlineEWC is more expensive than OnlineL2Reg because computing the Fisher also needs a virtual model learning step and storing the running sum of the previously stored matrices. 
The replay based methods store all raw data in memory.
ER is the most cheap because it does not need any local adaptation (i.e., virtual model update) for ranking. 
MIR and MaxLoss are almost equally expensive for ranking, and they both use the same lr and epochs of CFT for  virtual learning.





\end{document}